\documentclass{article}

\usepackage{arxiv}

\usepackage[utf8]{inputenc} 
\usepackage[T1]{fontenc}    
\usepackage{hyperref}       
\usepackage{url}            
\usepackage{booktabs}       
\usepackage{amsfonts}       
\usepackage{nicefrac}       
\usepackage{microtype}      
\usepackage{lipsum}
\usepackage{graphicx}
\graphicspath{ {./images/} }

\usepackage{amsmath}
\usepackage{times}
\usepackage{latexsym}
\usepackage[T1]{fontenc}
\usepackage[utf8]{inputenc}
\usepackage{microtype}
\usepackage{inconsolata}
\usepackage{algorithm}
\usepackage{algorithmic}

\usepackage[citestyle=authoryear,natbib=true,backend=bibtex]{biblatex}
\bibliography{bioncere.bib}

\title{BioNCERE: Non-Contrastive Enhancement For Relation Extraction In Biomedical Texts}

\author{
 Farshad Noravesh \\
  School of Information Technology\\
  Monash University \\
  \texttt{Farshad.Noravesh@monash.edu}
}

\begin{document}
\maketitle
\begin{abstract}
State-of-the-art models for relation extraction (RE) in the biomedical domain consider finetuning  BioBERT using classification, but they may suffer from the anisotropy problem. Contrastive learning methods can reduce this anisotropy phenomena, and also help to avoid class collapse in any classification problem. In the present paper, a new training method called biological non-contrastive relation extraction (BioNCERE) is introduced for relation extraction without using any named entity labels for training to reduce annotation costs. BioNCERE uses transfer learning and non-contrastive learning to avoid full or dimensional collapse as well as bypass overfitting. It resolves RE in three stages by leveraging transfer learning two times. By freezing the weights learned in previous stages in the proposed pipeline and by leveraging non-contrastive learning in the second stage, the model predicts relations without any knowledge of named entities. Experiments have been done on SemMedDB that are almost similar to State-of-the-art performance on RE without using the information of named entities. 
\end{abstract}

\section{Introduction}
Pretrained language models such as BioBert \citep{Lee2019} are essential for NLP tasks such as RE in the biomedical domain. However, contrastive learning could resolve some of the problems of these pre-trained models \citep{Jain2023} and many research papers have augmented their losses with contrastive loss and trained them jointly. \citep{Ozbulak2023} is a survey of all self-supervised learning methods that covers both contrastive and non-contrastive approaches and considers both generative and discriminative training methods. The term "non-contrastive" is emphasized to inform the reader that the model does not use any negative examples and only utilizes positive samples for training. Unsupervised relation extraction is promising since it does not require prior information on relation distribution and fixing the classes in advance. Thus it reduces the reliance on labeled data and can discover new relation types in raw text in a continual and online fashion. There are two ways to use contrastive learning namely supervised and self-supervised, although most papers focus on the latter one while the present paper focuses on the former.
\par
Many research papers augment the loss of a task with the contrastive loss for better data representation as will be described in this section, but contrastive learning has its challenges and issues. One of the problems of contrastive learning is that the number of negative samples should be large, but an increase in the negative samples would affect the false negative. An attempt to reduce this problem is augmenting three different losses that represent three different concerns namely the classical contrastive loss, the siamese style loss, and finally the prototypical cross-entropy \citep{Mo2022}. Although \citep{Mo2022} answers many concerns, the complexity of the K-means algorithm still is one of the major bottlenecks. \citep{Li2020} uses prototypes instead of comparison with all samples and considers prototypes as latent variables to perform expectation maximization. It also augments the prototypical loss with infoNCE loss \citep{Oord2018} to help bootstrap clustering.   
To make sure the model uses the full set of prototypes, \citep{Assran2022} augments the prototypical loss with the entropy of mean prediction across all anchor views. Most clustering methods are offline, but \citep{Caron2020} introduces an online algorithm in a way that the paradigm of contrastive instance learning is leveraged. One may treat each instance as a class to create a memory bank and the method of noise-contrastive estimation is used as an alternative to methods like negative sampling or hierarchical softmax which occurs in many other problems like word embedding \citep{Zhirong2018}. 
\par
There are non-contrastive methods such as "Bootstrap your own latent" (BYOL) that was introduced in \citep{Grill2020} which utilize a stop-gradient approach by making an asymmetric network and this prevents the model from collapsing. BYOL loss is used for non-contrastive learning in the present paper and it focuses on a supervised setting. Both \citep{Zhirong2018} and \citep{Grill2020} avoid using negative samples by leveraging different kinds of inductive bias. \citep{Chen2020} uses a symmetric loss similar to \citep{Grill2020} and avoids collapsing to constant. It shows an analysis that many algorithms like \citep{Li2020} could be considered as expectation maximization problems. For example, latent variables could be interpreted as cluster centers and these algorithms could be unified as an alternation framework that one fixes one network and update the other one as manifested in \citep{Li2020}. This periodic center update for clustering makes the algorithm slow which could be a bottleneck. \citep{Zhou2023} resolves this issue by combining the KL divergence loss between soft assignments and auxiliary distribution. Thus, \citep{Zhou2023} combines three losses namely the infoNCE loss which is an instance-wise contrastive loss, the cluster-wise contrastive loss, and the KL divergence loss.
\par
An alternative approach to using stop gradient to avoid collapsing to constant is appreciating the idea that the hierarchy of classes could provide an inductive bias as is described in \citep{Liu2022}, where they used hierarchical propagation clustering to generate clusters of different granularities and the momentum encoder smoothes the updates which are similar to the idea of stop gradient in \citep{Grill2020} or \citep{Chen2020}. The present paper tries to resolve the issue of collapse which is discussed in \citep{Li2022} and it was found experimentally that SimSiam could fail if the model is too small relative to the dataset that is being used. Although \citep{Zhang-Chaoning2022} introduces vector decomposition to analyze the collapse in simSiam, the main underlying reasons for avoiding collapse in these networks are still not known. \citep{Shi2020} introduces RAFT and proves that it is equivalent to BYOL under certain conditions. 
\par
Some methods for non-contrastive learning are very easy to implement, but the underlying theoretical analysis of them is less known such as the Barlow twins model which minimizes feature redundancy while maximizing invariance to common corruptions. Although the implementation of Barlow Twins in \citep{Bandara2023} for contrastive learning in \citep{Zbontar2021} looks very simple, the analysis is not yet understood. The simplicity comes from the fact that Barlow Twins does not require a large batch size or stop gradient. In contrast to methods like SimSiam or BYOL, Barlow Twins does not need predictor networks or momentum encoders. \citep{Tian2020} show the importance of view selection and argue that mutual information between the views should be minimized and a good way to do it is by proper data augmentation. 
\par
Most articles on Contrastive losses are focused on self-supervised framework but few articles consider supervised setting of contrastive learning such as \citep{Reimers2019} that is done for sentence embedding over NLI dataset which introduces SBERT model to have a much better semantic representation in comparison with traditional TF-IDF methods that only use surface forms of words and neglect the context and semantics. Although SBERT is a great attempt, in the real world labeled data are expensive. Thus, \citep{Zhang2021} leverages data augmentation in an unsupervised way. \citep{Zhang2021} avoids collapsing to constant by leveraging the idea of \citep{Grill2020} that utilizes stop-gradient. \citep{Zhang2023} uses Siamese representation for relation extraction. Using contrastive learning for relation extraction is not new and is done in \citep{Su2021}. However, \citep{Su2021} uses data augmentation by removing random words in a sentence and is unsupervised. The present paper is done in a supervised setting since a huge amount of labeled data is available in SemMedDB \citep{Kilicoglu2012} which is a repository of semantic predications (subject–predicate–object triples) extracted from the entire set of PubMed citations. Thus, positive samples are naturally those samples that have the same relations. A preprocessing is done to gather all sentences that have the same relations and the datapoints inside each minibatch have the same relations. Contrastive learning was first used in an unsupervised setting but supervised approaches also exist such as \citep{Khosla2020}. Many articles have used contrastive learning for relation extraction such as \citep{Yuan2023}. It is a common practice to augment classification loss based on cross-entropy with the contrastive loss or other losses to satisfy different objectives simultaneously as is implemented in \citep{Yasar2023} in the setting of continual learning or zero-shot learning in \citep{Khalifa2023}. \citep{Sundareswaran2021} combines the objective of clustering with contrastive loss or \citep{Li2023} leverages contrastive learning for better meta learning.
\citep{Chen2022} introduces SupCon and uses transfer learning for supervised contrastive learning and proves that adding a weighted class-conditional InfoNCE loss to SupCon controls the degree of spread. \citep{Zhang-Guojun2022} focuses on choosing proper f-divergences to characterize better alignment and uniformity losses. 
\par
The latest version of SemMedDB is used in the present paper which was first introduced in \citep{Kilicoglu2012} but it has been continuously updated. 
The state-of-the-art model for RE in the SemMedDB dataset is " Biomedical Predicate Relation-extraction with Entity-filtering by PKDE4J " (BioPrep) \citep{Hong2021}, and the present work compares F1 scores with it although BioPrep leverages named entities and grouping for preprocessing while the present model directly approaches the problem.
\par
The following are four major contributions of the present paper:
\begin{itemize}
\item {Augmenting contrastive loss to avoid producing anisotropy distribution of pretrained and finetuned model.
 }
\item {Using BYOL for NLP and in a supervised manner in contrast to the original paper which was for computer vision and was fully self-supervised.}
\item {Proving experimentally that the state-of-the-art model’s accuracy can be achieved with only relation(predicate) information and the entity annotations are not needed.}
\item {Introducing a new training method instead of traditional joint training for relation extraction of biomedical texts that is scalable and can be analyzed in three independent stages: 
finetune BioBert then fine-tune contrastive network and finally classification for relation extraction which is in harmony with the “separation of concerns design principle” in software development.}
\end{itemize}

\section{Modeling}
The main task in the present paper is RE, but non-contrastive learning is leveraged for better representation of sentences. 
\subsection{Problem Formulation}
\begin{figure}[h!]
\centering
 \includegraphics[width=8cm,height=8cm]{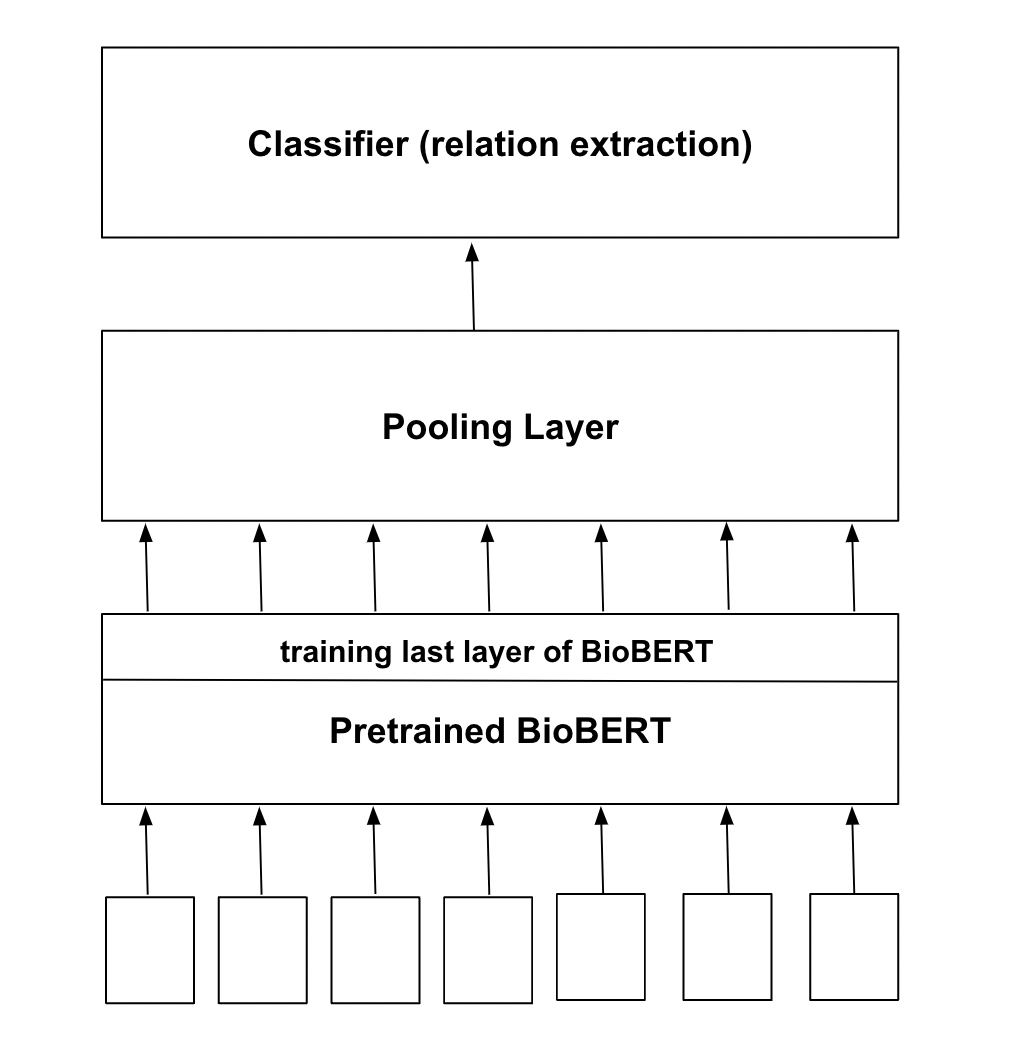}
 \caption{finetuning BioBERT \label{fig.finetuning}}
\end{figure}

The proposed method, BioNCERE is composed of three modules as illustrated in Figure~\ref{pipeline} which shows how the input comes from the SemMedDB database and then three modules process this input and finally the output of the algorithm is the predicate(relations) information. 
The first one is called the representation module and involves finetuning a pre-trained BioBERT model and tuning it for RE as is shown in Figure~\ref{fig.finetuning}. Since pre-trained models are anisotropic, this step alone does not provide good performance for the RE task due to a lack of appropriate sentence representation. The weights of this model are frozen after training and are passed over to the next module. 
Given a pre-trained BioBERT model, a nonlinear layer composed of a linear network and a nonlinear activation is used for finetuning. During this process, all weights of the original BioBERT model are frozen except the last layer. Figure~\ref{fig.finetuning} shows the finetuning architecture of BioBERT. All the layers of BioBERT are frozen except the last layer. The hidden representation of the last layer is given to a pooling mechanism to create a vector which is then fed into a nonlinear layer of the classifier. Many pre-trained models based on transformer architecture suffer from anisotropy \citep{Jain2023}. Experiments in \citep{Ethayarajh2019} show that In all layers of ELMo, BERT, and GPT-2, the contextualized word representations of all words are not isotropic and they are located in a narrow cone in the vector space. A similar problem is called token uniformity which is explained in \citep{Yan2022}. To avoid this collapse to certain outlier dimensions, the present work stops the finetuning of BioBert at early stages so that the non-contrastive network does not receive vectors that are too sensitive in some dimensions. 
\par
The second module is a non-contrastive network as is shown in Figure~\ref{fig.sentenceRep}, and the goal is to enhance the representation of the previous module using a non-contrastive loss. Note that the original non-contrastive network of BYOL was introduced in the framework of self-supervised learning but the present paper only uses supervised non-contrastive learning since there are a huge number of sentences available in SemMedDB and there is no need for data augmentation in the self-supervised setting. A simple preprocessing step is done to distribute the positive samples in two main pathways namely the online path and the target path. With the same analogy, the weights of the second module are frozen after training to be used for the third module. The third module is called the classifier which receives the representation of the second module to classify the sentence into 28 classes.
\par
The third module is shown on the left side of Figure~\ref{fig.sentenceRep}. By the same analogy and justification for the first module, transferring the weights from the non-contrastive network to the final classification layer is implemented since any contrastive loss may still collapse slightly to certain dimensions. The second and third modules can be trained jointly, but this leads to a complex training procedure that may easily produce an overfitting issue. Thus, the suggested training procedure is shown in Algorithm~\ref{alg:bioncere}.

\subsection{Contrastive Enhancement}
In contrast to \citep{Zhang2021} which uses data augmentation, the present work is using supervised learning since many labeled data are available in SemMedDB and there is no necessity for data augmentation. 
\begin{figure}[h!]
\centering
 \includegraphics[width=8cm,height=8cm]{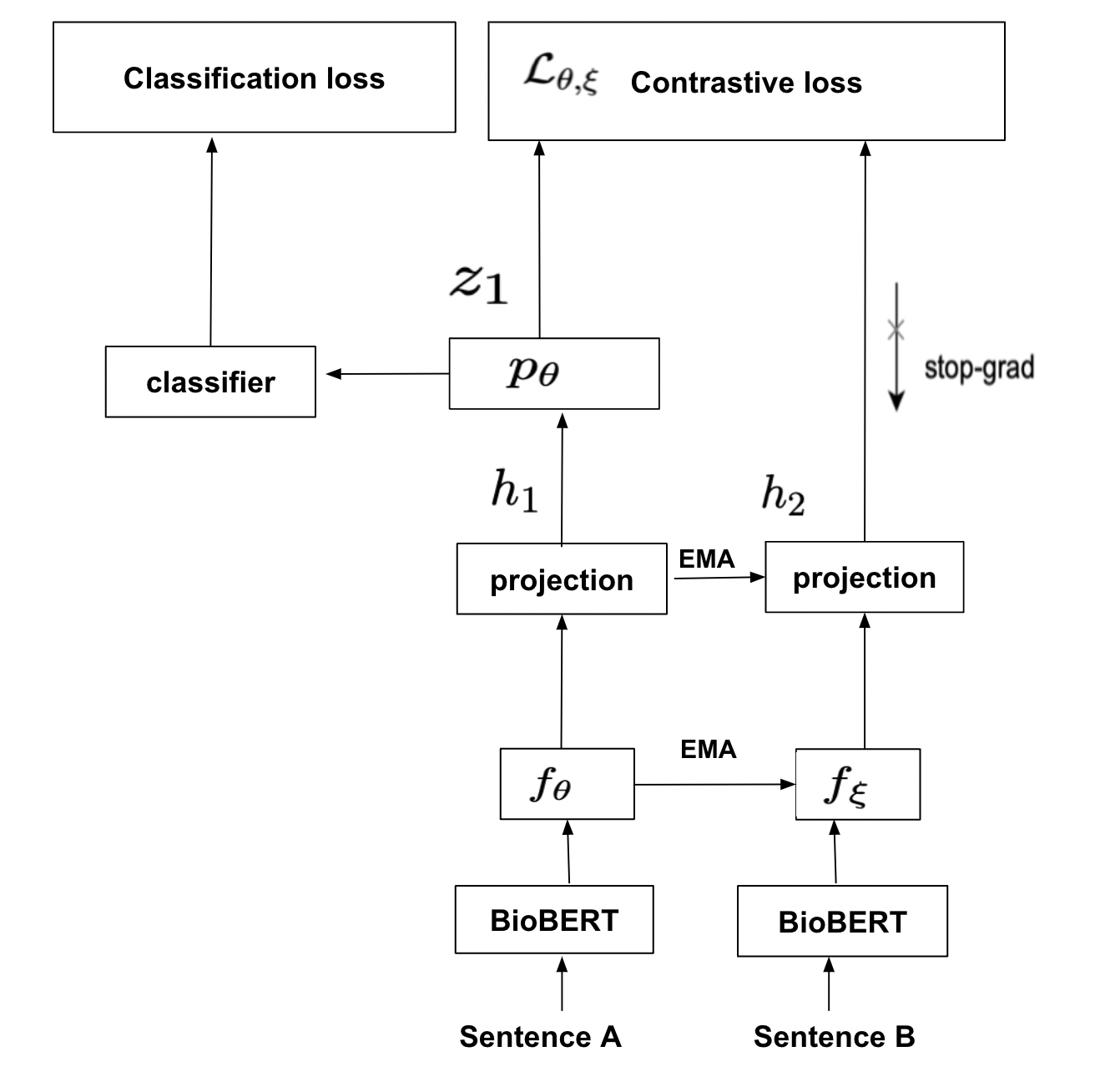}
 \caption{architecture of BioNCERE \label{fig.sentenceRep}}
\end{figure}
Figure~\ref{fig.sentenceRep} shows the proposed model which contains two networks. 
Using the same naming convention in the BYOL paper, the first network is called the online network which contains the encoder $f_{\theta}$, projection, and the predictor $p_{\theta}$. The second network is called the target network which contains only an encoder $f_{\xi}$ and projection but lacks a predictor. The goal is to minimize their negative cosine similarity which is defined by:
\begin{equation}
D_{\theta,\xi}(z_{1},h_{2}) = - <\frac{z_1}{||z_1||},\frac{h_2}{||h_2||}>
\end{equation} 
An asymmetric loss is defined between two sentences as follows:
\begin{equation}
\mathcal{L}_{\theta,\xi} = \frac{1}{2}D_{\theta,\xi}(z_{1},h_{2}) + \frac{1}{2}D_{\theta,\xi}(\tilde{z}_{2},\tilde{h}_{1})
\end{equation}
where $\tilde{z}_{2} = p_{\theta}(f_{\theta}(x_{2}))$ and $\tilde{h}_{1} = f_{\xi}(x_{1})$ 
We follow the stop gradient idea of \citep{Grill2020} and update the weights as follows:
\begin{equation}
\begin{split}
\theta_{t} &= \theta_{t-1} + \nabla_{\theta} \mathcal{L}_{\theta,\xi} \\
\xi_{t} &= \delta \xi_{t-1} + (1-\delta)\theta_{t}
\end{split}
\end{equation}
\par
The encoder is a BioBERT network that was first introduced in \citep{Lee2019}. The Encoder uses a BioBert to represent the sentence. A simple pooling mechanism is used to use the final hidden state as a representation of the whole sentence. 
The following cross-entropy loss is used for RE :
\begin{equation}
\mathcal{L}_{cls}(y,\hat{y}) = -\frac{1}{N} \sum_{i=1}^{N} y_{i}^{T} \log{\hat{y}_{i}}
\end{equation}
\par
\begin{figure*}[h!]
\centering
 \includegraphics[width=16cm,height=4cm]{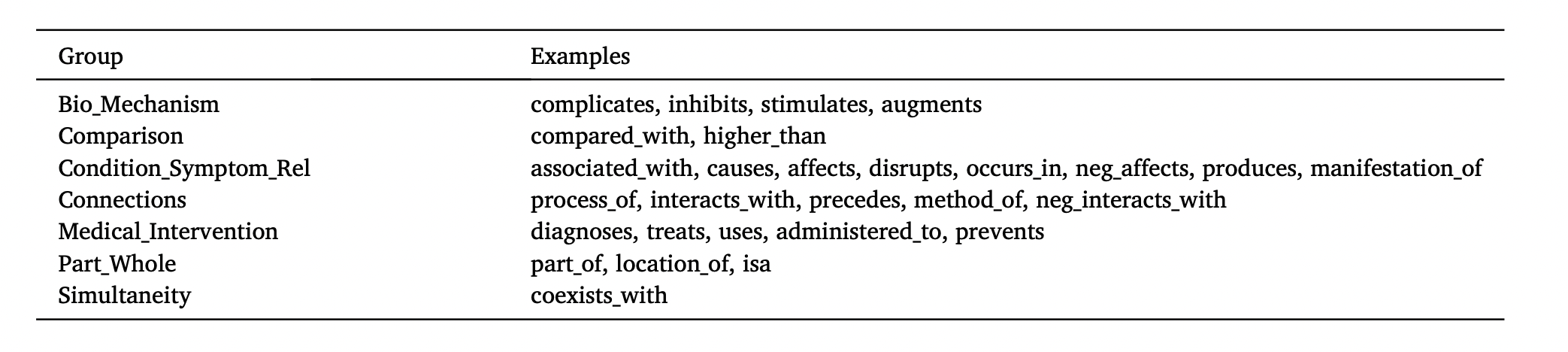}
 \caption{28 predicates are grouped into 7 categories \label{fig.28predicates}}
\end{figure*}

The original SemMedDB dataset contains 67 predicates but to compare results with state of the art model which is BioPrep, only 28 predicates out of 66 are chosen as illustrated in Figure~\ref{fig.28predicates}. By leveraging the enhanced representation of module 2, a linear layer is designed to classify into 28 predicates.

\begin{algorithm}

  \begin{algorithmic}
   
    \STATE Input : (sentence, relation) from SemMedDB \\
    \STATE
    \STATE 1: connecting BioBERT to a single linear layer \\
    \STATE 2: FineTuning frozen BioBERT except for its last layer and the linear layer using cross-entropy loss \\
    \STATE 3: Freezing the learned weights in previous steps \\
    \STATE 4: arranging supervised data in pairs as augmentation \\
    \STATE 5: Learning using non-contrastive loss \\
    \STATE 6: Freezing the weights of non-contrastive network \\
    \STATE 7: learning the classifier using cross-entropy loss \\
    \STATE
    \STATE Output: predicate prediction
    \caption{BioNCERE algorithm}
    \label{alg:bioncere}
  \end{algorithmic}

\end{algorithm}
\begin{table*}[h!]
\centering
\begin{tabular}  {
p{2.7cm}p{1.5cm}p{1.5cm}p{1.5cm}p{1.5cm}p{1.5cm}p{1.5cm}p{1.5cm} }
\hline
\textbf{Model} & \textbf{BioNCere} & \textbf{BioNCere} & \textbf{BioNCere} & \textbf{BioPrep} & \textbf{BioPrep} & \textbf{BioPrep}  \\
\hline
\textbf{batch} & \textbf{64} & \textbf{64} & \textbf{64} & \textbf{64} & \textbf{64} & \textbf{64}  \\
\hline
\textbf{} & \textbf{Precision} & \textbf{Recall} & \textbf{F1 Score} & \textbf{Precision} & \textbf{Recall} & \textbf{F1 Score} \\
\hline
complicates & 1.00 & 1.00 & 1.00 & 0.91 & 0.88 & 0.90\\
\hline
inhibits\textunderscore than         & 0.67 & 0.50 & 0.57 & 0.82 & 0.83 & 0.83 \\
\hline
stimulates & 0.75 & 0.62 & 0.71 & 0.89 & 0.85 & 0.87 \\
\hline
augments & 0.83 & 0.62 & 0.71 & 0.80 & 0.85 & 0.83 \\
\hline
compared\textunderscore with & 0.67 & 1.00 & 0.80 & 0.89 & 0.96 & 0.92 \\
\hline
higher\textunderscore than & 0.67 & 1.00 & 0.80 & 0.83 & 1.00 & 0.91 \\
\hline
associated\textunderscore with & 1.00 & 1.00 & 1.00 & 0.82 & 0.81 & 0.81 \\
\hline
causes & 1.00 & 1.00 & 1.00 & 0.88 & 0.90 & 0.89 \\
\hline
affects & 0.67 & 0.67 & 0.67 & 0.82 & 0.85 & 0.83 \\
\hline
disrupts & 0.50 & 0.50 & 0.50 & 0.82 & 0.81 & 0.82 \\
\hline
occurs\textunderscore in & 0.67 & 1.00 & 0.80 & 0.63 & 0.75 & 0.68 \\
\hline
neg\textunderscore affects & 0.33 & 1.00 & 0.50 & 0.94 & 0.85 & 0.89 \\
\hline
produces & 1.00 & 0.50 & 0.67 & 0.75 & 0.85 & 0.80 \\
\hline
manifestation\textunderscore of & 1.00 & 0.50 & 0.67 & 0.65 & 0.87 & 0.74 \\
\hline
process\textunderscore of & 0.75 & 0.60 & 0.67 & 0.87 & 0.87 & 0.87 \\
\hline
interacts\textunderscore with & 1.00 & 1.00 & 1.00 & 0.86 & 0.88 & 0.87 \\
\hline
precedes & 1.00 & 1.00 & 1.00 & 0.89 & 0.71 & 0.79 \\
\hline
method\textunderscore of & 1.00 & 1.00 & 1.00 & 0.67 & 0.70 & 0.68 \\
\hline
neg\textunderscore interacts\textunderscore with & 0.67 & 1.00 & 0.80 & 0.88 & 0.82 & 0.85 \\
\hline
diagnoses & 1.00 & 0.75 & 0.86 & 0.80 & 0.84 & 0.81 \\
\hline
treats & 0.75 & 1.00 & 0.86 & 0.85 & 0.85 & 0.85 \\
\hline
uses & 1.00 & 0.4 & 0.57 & 0.84 & 0.80 & 0.82 \\
\hline
administered\textunderscore to & 1.00 & 1.00 & 1.00 & 0.85 & 0.90 & 0.87 \\
\hline
prevents & 1.00 & 1.00 & 1.00 & 0.79 & 0.82 & 0.81 \\
\hline
part\textunderscore of & 1.00 & 1.00 & 1.00 & 0.86 & 0.85 & 0.86 \\
\hline
location\textunderscore of & 1.00 & 0.5 & 0.67 & 0.85 & 0.83 & 0.84 \\
\hline
isa & 0.33 & 1.00 & 0.50 & 0.82 & 0.74 & 0.78 \\
\hline
coexists\textunderscore with & 0.5 & 1.00 & 0.67 & 0.82 & 0.85 & 0.84 \\
\hline
average & 0.848 & 0.827 & 0.785 & 0.825 & 0.84 & 0.800
\end{tabular}
\caption{precision, recall and F1score of BioNCere and BioPrep for 28 predicates}
\label{tab:results_small_batch}
\end{table*}

\begin{table*}[h!]
\centering
\begin{tabular}{ p{2.7cm}p{1.5cm}p{1.5cm}p{1.5cm}p{1.5cm}p{1.5cm}p{1.5cm}p{1.5cm} }
\hline
\textbf{Model} & \textbf{BioNCere} & \textbf{BioNCere} & \textbf{BioNCere} & \textbf{BioNCere} & \textbf{BioNCere} & \textbf{BioNCere}  \\
\hline
\textbf{batch} & \textbf{128} & \textbf{128} & \textbf{128} & \textbf{256} & \textbf{256} & \textbf{256}  \\
\hline
\textbf{} & \textbf{Precision} & \textbf{Recall} & \textbf{F1 Score} & \textbf{Precision} & \textbf{Recall} & \textbf{F1 Score} \\
\hline
complicates & 0.75 & 1.00 & 0.86 & 0.89 & 1.00 & 0.94\\
\hline
inhibits\textunderscore than         & 0.33 & 0.33 & 0.33 & 0.44 & 0.40 & 0.42 \\
\hline
stimulates & 0.44 & 1.00 & 0.62 & 0.64 & 0.56 & 0.60 \\
\hline
augments & 0.67 & 0.80 & 0.73 & 0.60 & 0.46 & 0.52 \\
\hline
compared\textunderscore with & 0.57 & 1.00 & 0.73 & 0.57 & 0.73 & 0.64 \\
\hline
higher\textunderscore than & 0.75 & 1.00 & 0.86 & 0.60 & 1.00 & 0.75 \\
\hline
associated\textunderscore with & 0.75 & 0.60 & 0.67 & 0.57 & 0.80 & 0.67 \\
\hline
causes & 1.00 & 0.67 & 1.00 & 0.78 & 0.78 & 0.78 \\
\hline
affects & 0.40 & 0.50 & 0.44 & 0.67 & 0.40 & 0.50 \\
\hline
disrupts & 0.33 & 0.60 & 0.43 & 0.86 & 1.00 & 0.92 \\
\hline
occurs\textunderscore in & 0.86 & 1.00 & 0.92 & 0.86 & 0.86 & 0.86 \\
\hline
neg\textunderscore affects & 0.67 & 1.00 & 0.80 & 0.82 & 0.90 & 0.86 \\
\hline
produces & 0.57 & 1.00 & 0.73 & 0.53 & 0.80 & 0.84 \\
\hline
manifestation\textunderscore of & 1.00 & 0.80 & 0.89 & 0.88 & 0.78 & 0.82 \\
\hline
process\textunderscore of & 0.33 & 0.25 & 0.29 & 0.5 & 0.5 & 0.5 \\
\hline
interacts\textunderscore with & 0.67 & 0.40 & 0.50 & 0.62 & 0.45& 0.53 \\
\hline
precedes & 0.50 & 0.67 & 0.57 & 1.00 & 1.00 & 1.00 \\
\hline
method\textunderscore of & 0.40 & 0.50 & 0.44 & 0.89 & 0.80 & 0.84 \\
\hline
neg\textunderscore interacts\textunderscore with & 0.60 & 1.00 & 0.75 & 0.80 & 0.73 & 0.76 \\
\hline
diagnoses & 0.75 & 0.60 & 0.67 & 1.00 & 0.70 & 0.82 \\
\hline
treats & 0.60 & 0.38 & 0.46 & 0.62 & 0.73 & 0.67\\
\hline
uses & 0.80 & 0.44 & 0.57 & 0.80 & 0.57 & 0.67 \\
\hline
administered\textunderscore to & 0.86 & 0.86 & 0.86 & 0.80 & 0.89 & 0.84 \\
\hline
prevents & 1.00 & 1.00 & 1.00 & 0.71 & 0.83 & 0.77 \\
\hline
part\textunderscore of & 0.50 & 0.25 & 0.33 & 0.25 & 0.25 & 0.25 \\
\hline
location\textunderscore of & 0.50 & 0.17 & 0.25 & 0.50 & 0.29 & 0.36 \\
\hline
isa & 0.25 & 0.20 & 0.22& 0.67 & 0.33 & 0.44 \\
\hline
coexists\textunderscore with & 1.00 & 0.67 & 0.80 & 0.44 & 0.40 & 0.42 \\
\hline
average & 0.637 & 0.667 & 0.632 & 0.689 & 0.676 & 0.648
\end{tabular}
\caption{precision, recall and F1score of BioNCERE for different batch sizes}
\label{tab:results_big_batch}
\end{table*}

\subsection{Training}
\begin{figure*}[h!]
\centering
 \includegraphics[width=15cm,height=4cm]{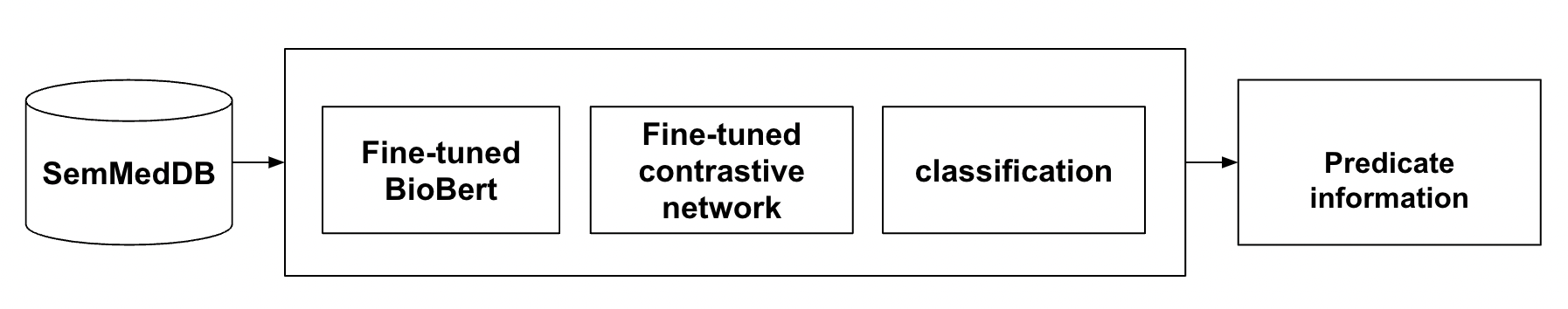}
 \caption{pipeline of BioNCERE \label{pipeline}}
\end{figure*}

Figure~\ref{pipeline} shows the pipeline of BioNCERE. Please note that in BioNCERE, information on entities is not used and relations are predicted directly, while in BioPrep entity extraction and some filtration mechanisms are a part of the pipeline before the final RE task.
\par
The non-contrastive loss is combined with the classification loss to simultaneously enhance the primary representation as well as classify the sentence into 28 predicates. The total loss is as follows:
\begin{equation}\label{eq:total_loss}
\mathcal{L}_{tot} = \mathcal{L}_{cls} + \lambda \mathcal{L}_{cont}
\end{equation}
where $\lambda$ in Equation~\eqref{eq:total_loss} is a hyper-parameter that indicates the tradeoff between enhancing representation and classification.
There are many methods for training. The first approach is jointly training both the contrastive loss and the classification loss as is shown in \eqref{eq:total_loss}, but this approach has one major issue. The issue is that during the training period, the backpropagation from the classifier may adversely affect the attempts of the non-contrastive network, and therefore the model may either fully collapse or suffer from dimensional collapse and the model reaches a local minimum. Thus, a very complex and dynamic mechanism should be designed to dynamically update the $\lambda$ instead of fixing it to avoid overfitting.
In contrast to the joint training paradigm, the present work leverages transfer learning. After initial finetuning of the BioBert for RE task, the non-contrastive network is trained and then the pre-trained non-contrastive network is frozen to be used for the third module which is final finetuning by classification to get the final predicate probabilities. This leads to better separability of different concerns and objectives like non-contrastive loss and classification loss. It also encourages the reusability of different subtasks that could be used for other downstream tasks such as few-shot learning.

\section{Experiments}

\subsection{Experimental Data And Setup}
SemMedDB database is used to collect sentences and their relations. Some SQL queries are written as a preliminary step to fetch necessary data and to create two columns, namely sentences and their corresponding predicates. It has more than 119.1 M sentences from PubMed citations. The database has many tables including more than 12.9 Million predication and over 1.3 Million concepts from 21 Million citations.
There exist very few works that have done RE on SemMedDB such as \citep{Hong2021}. They evaluated their experiments by showing the precision, recall, and F1 scores of individual predicates. 
Figure~\ref{fig.28predicates} shows that they picked only 28 predicates out of 67 predicates and grouped them into seven categories as illustrated in Figure~\ref{fig.28predicates}. Data for the present work is limited to 56000 sentences and their corresponding relations(28 predicates out of 67) which are split equally into train, evaluation, and test datasets. 

\subsection{Results And Discussions}
\citep{Chen2020} showed that non-contrastive learning benefits from larger batch sizes and more training steps compared to traditional supervised learning where batch sizes may not be a very sensitive variable in experiments. Thus, different batch sizes are investigated in the present work. During transfer learning from pre-trained BioBERT to finetuned relation extraction model, we noticed the model overfits if all weights of the original BioBERT model are frozen. Thus, the weights of the final layer of BioBERT are allowed to be learnable and are involved in the backpropagation of error. After a preprocessing step, two minibatches namely $batch\textunderscore a$ and $batch\textunderscore b$ are created so that sentences of $batch\textunderscore a$ have their corresponding positive sample in $batch \textunderscore b$ but the sentences of each batch belongs to different relation categories. This step is essential in any contrastive algorithm.
\par
 Although different ways to create and treat batches are introduced in \citep{Cho2023}, the present experiments are focused on the traditional way for simplicity and better clarity of fair comparison. By comparing Table~\ref{tab:results_big_batch} with Table~\ref{tab:results_small_batch}, it is observed that training with high batch sizes such as 128 and 256 could underfit. Any batch size below 64 could lead to overfitting since there are not enough examples for each class inside each batch and the model tends to attend to particular classes and ignore the rest. It takes 20 hours for a model with 1 Million parameters to be trained for 15 epochs with a single GPU. To increase model complexity, one may need to unfreeze more layers of BioBert which could lead to higher computations and more RAM and GPU resources. 

\section{Conclusion}
A three-stage modeling is designed by leveraging transfer learning two times. In the first stage, the pre-trained BioBert is fine-tuned by the RE task. Then the learned parameters are fed to a non-contrastive network. The non-contrastive network is trained separately and provides a suitable sentence representation. The parameters of the non-contrastive network are then frozen and fed to the final classification module. These three stages make the algorithm very flexible and can be done by three different teams of developers and respect the "separation of concern principle" in software development. The results show that higher batch sizes tend to underfit while batch sizes below 64 could lead to overfitting and the appropriate number of batch sizes is around 64.  

\section*{Limitations}
Although the present work could not outperform the State-of-the-art models for relation extraction (RE) in the biomedical domain for the SemMedDB database, it has the advantage of not requiring named entity information during the training process.


\printbibliography

\end{document}